\documentclass[10pt,twocolumn,letterpaper]{article}

\usepackage[pagenumbers]{wacv} % To force page numbers, e.g. for an arXiv version

\usepackage[utf8]{inputenc}
\usepackage[T1]{fontenc}
\usepackage{times}
\usepackage{graphicx}
\usepackage{amsmath,amssymb}
\usepackage{booktabs}
\usepackage{url}

\definecolor{wacvblue}{rgb}{0.21,0.49,0.74}
\usepackage[breaklinks,colorlinks,allcolors=wacvblue]{hyperref}
\usepackage{multirow}
\def\wacvPaperID{*****} % *** Enter the WACV Paper ID here
\def\confName{WACV}
\def\confYear{2026}

\newcommand{\model}{OMNI-Dent}

\title{\model: Towards an Accessible and Explainable AI Framework for Automated Dental Diagnosis}

\author{
Leeje Jang$^{1}$ \quad
Yao-Yi Chiang$^{1}$ \quad
Angela M. Hastings$^{1}$ \quad
Patimaporn Pungchanchaikul$^{2}$ \\
Martha B. Lucas$^{1}$ \quad
Emily C. Schultz$^{3}$ \quad
Jeffrey P. Louie$^{1}$ \quad
Mohamed Estai$^{4}$ \\
Wen-Chen Wang$^{5}$ \quad
Ryan H.L. Ip$^{6}$ \quad
Boyen Huang$^{1}$ \\
$^{1}$University of Minnesota, United States \quad
$^{2}$Khon Kaen University, Thailand \\
$^{3}$Minnesota State University, Mankato, United States \quad
$^{4}$The University of Western Australia, Australia \\
$^{5}$Kaohsiung Medical University, Taiwan \quad
$^{6}$Auckland University of Technology, New Zealand \\}
\begin{document}
\maketitle

\begin{abstract}
% % 200-250 words
Accurate dental diagnosis is essential for oral healthcare, yet many individuals lack access to timely professional evaluation. Existing AI-based methods primarily treat diagnosis as a visual pattern recognition task and do not reflect the structured clinical reasoning used by dental professionals. These approaches also require large amounts of expert-annotated data and often struggle to generalize across diverse real-world imaging conditions. To address these limitations, we present \model, a data-efficient and explainable diagnostic framework that incorporates clinical reasoning principles into a Vision-Language Model (VLM)-based pipeline. The framework operates on multi-view smartphone photographs, embeds diagnostic heuristics from dental experts, and guides a general-purpose VLM to perform tooth-level evaluation without dental-specific fine-tuning of the VLM. By utilizing the VLM's existing visual-linguistic capabilities, \model\ aims to support diagnostic assessment in settings where curated clinical imaging is unavailable. We design \model\ as an early-stage assistive tool to help users identify potential abnormalities and determine when professional evaluation may be needed, thereby offering a practical option for individuals with limited access to in-person care.
\end{abstract}

\section{Introduction}
\label{sec:intro}  % This corresponds to your ``2. Introduction''

Oral health plays an important role in maintaining quality of life across the human lifespan, yet many individuals, particularly those in underserved or rural communities~\cite{rahman2024dental,peres2019oral}, struggle to access timely and reliable dental diagnosis. Limited availability of dental professionals often leads to delayed treatment and preventable deterioration in oral health. Prior work on teledentistry~\cite{azimi2025teledentistry, huang2024mobile,estai2022mobile,estai2017comparison,estai2016efficacy,estai2016proof,schultz2025review} provides remote diagnostic support to improve accessibility; however, limited expert availability and the resources required for remote evaluation still remain ongoing challenges. 
% At the same time, smartphones are widely accessible and provide a practical means for low-cost, scalable oral health assessment through dental photographs.

Recent advances in artificial intelligence (AI) motivate the development of automated dental diagnostic systems. Existing approaches~\cite{wang2025application,abbott2025artificial,dan2025artifact,santos2025unique,zhang2022development} typically rely on deep learning models trained to detect or classify dental conditions from clinical imaging modalities such as radiographs or intraoral photographs taken with professional devices. Although effective in controlled settings, these methods require large expert-labeled datasets in which clinicians manually annotate tooth locations and conditions, making the annotation process labor-intensive and difficult to scale. Model performance is also sensitive to visual similarity between the training data and deployment environments, limiting applicability across the wide range of real-world dental imaging conditions. A fundamental limitation of current AI-based dental systems is that they treat diagnosis largely as a visual pattern recognition task. In practice, however, dental professionals use structured clinical reasoning that involves comparing neighboring and contralateral teeth, considering multi-surface morphology, and interpreting structural changes such as wear, erosion, or cavities in the context of the full dentition. Existing AI-based methods do not encode these diagnostic heuristics, so they fail to reflect how clinicians reason through a diagnosis and result in a lack of interpretability.

The emergence of Generative AI (GenAI), particularly Vision-Language Models (VLMs) ~\cite{chen2024internvl,wang2025internvl3,lu2024ovis,bai2023qwen,achiam2023gpt} trained on large-scale image-text pairs, introduce new possibilities for advancing automated dental diagnosis. These models learn from various image-understanding tasks (e.g., image captioning, visual question answering, object detection) across different domains. However, applying VLMs to dentistry ~\cite{wu2025chatios,du2024prompting} typically require  additional domain-specific datasets annotated by experts, leading to substantial labeling costs and systems that often overfit to narrow clinical imaging conditions. Recent work also introduces VLMs~\cite{sellergren2025medgemma} and benchmark datasets tailored for medical imaging~\cite{haotowards}, but these efforts continue to focus on professionally captured modalities such as radiographs or intraoral photographs obtained with specialized devices rather than accessible smartphone-based images. These limitations highlight the need for an approach that minimizes reliance on dental-specific annotations and controlled clinical imaging environments while effectively leveraging the existing capabilities of modern VLMs.

To address these challenges, we propose \model, an explainable and data-efficient diagnostic framework designed to emulate the reasoning process of dental clinicians and to serve as an early, first-line assistive tool for identifying subtle abnormalities. \model\ operates on multi-view smartphone photographs consisting of frontal, upper occlusal, and lower occlusal views, making it suitable for use outside specialized clinical imaging environments. A key component of the framework is a clinical reasoning module that directs a general-purpose VLM to follow expert-defined diagnostic steps, enabling tooth-level evaluation without any fine-tuning of the VLM. The goal is to provide accessible, initial screening support that helps individuals detect potential dental issues at an early stage and determine when professional care is warranted. We examine how a state-of-the-art VLM pretrained on broad image-text corpora can be guided to follow clinician-like diagnostic reasoning. By leveraging relational cues within the dentition and embedding expert heuristics, \model\ performs tooth-level diagnosis while retaining the VLM's general visual capabilities. Operating directly on smartphone photographs rather than specialized clinical images enables early at-home assessment for individuals who face barriers to in-person evaluation, including those in underserved or geographically isolated communities.

We summarize the main contributions of this work as follows:
\begin{itemize}
\item We present \model, an explainable and data-efficient diagnostic framework that performs tooth-level diagnosis from smartphone photographs using a general-purpose VLM, without any fine-tuning.

\item We propose a clinical reasoning module that guides the VLM through structured diagnostic steps inspired by how dental professionals evaluate teeth, using explicit visual cues to support reliable tooth-level interpretation.

\item We demonstrate how state-of-the-art VLMs perform dental imaging tasks in both zero-shot and few-shot in-context learning (ICL) settings, achieving strong quantitative performance across multiple diagnostic categories.

\item We highlight the potential of \model\ as an early at-home screening tool that operates on smartphone photographs, offering accessible diagnostic support for individuals who face barriers to in-person dental care.
\end{itemize}

\section{Related Work}
\label{sec:related}   % This corresponds to your ``3. Related Work''
\subsection{Accessible dental diagnosis}
Oral health represents an essential component of overall health and quality of life, yet access to dental care remains uneven across populations. Prior work shows that individuals in underserved or geographically isolated communities ~\cite{rahman2024dental} continue to face substantial barriers in obtaining timely professional evaluation. Existing studies ~\cite{azimi2025teledentistry, huang2024mobile,estai2022mobile,estai2017comparison,estai2016efficacy,estai2016proof,schultz2025review} therefore explore teledentistry as a remote diagnosis framework to facilitate online diagnosis and consultation for individuals who cannot easily reach a clinic. Specifically, recent work~\cite{huang2024mobile} evaluates the diagnostic accuracy of smartphone photographs for traumatic dental injuries (TDI), demonstrating strong potential for expanding remote dental care through smartphone-based assessment. Although existing studies report encouraging diagnostic performance for teledentistry-based assessments, these services still depend on human experts to interpret remotely submitted information. This dependence introduces practical constraints because expert availability and the resources required for remote evaluation are not reliably accessible in a timely manner.

\subsection{AI for Dental Diagnosis}
AI-based methods increasingly serve dental diagnosis, with prior work~\cite{wang2025application, abbott2025artificial, dan2025artifact, santos2025unique, zhang2022development} primarily adopting convolutional neural networks and other image-driven architectures to identify conditions in radiographs~\cite{dan2025artifact} and clinical photographs~\cite{zhang2022development}. Training these models typically relies on supervised learning with dental images, which requires large amounts of task-specific data that share similar visual appearances and diagnosis codes, along with expert-annotated labels that demand substantial human effort. Although these approaches learn visual patterns associated with disease, they still treat diagnosis primarily as a pattern recognition task, which makes it difficult to understand how the model reaches a particular decision and results in limited interpretability.

% These models generally operate by discovering visual patterns associated with each condition. However, they rely on large labeled datasets and do not explicitly encode the diagnostic reasoning processes used in clinical practice.
% As a result, existing work provide limited interpretability and do not reflect the structured decision-making processes used by human experts. Moreover, their performance typically depends on large, carefully curated datasets, which are costly and time-consuming to assemble in domains that require expert-level annotation.

\subsection{VLM for Dental Diagnosis}

Vision-language models, particularly Large-VLMs that combine a visual encoder with an LLM-style decoder to generate natural language descriptions of images (e.g., GPT4~\cite{achiam2023gpt}, OVIS~\cite{lu2024ovis}, Qwen~\cite{bai2023qwen}, and the InternVL series~\cite{chen2024internvl,wang2025internvl3}), demonstrate strong capabilities across general visual-linguistic tasks and support the integration of visual information with textual interpretation and reasoning. VLMs learn from large amounts of image-text pairs that include tasks such as image captioning and visual question answering (VQA). In addition, in-context learning (ICL) enables the models to adapt from general-purpose training to domain-specific tasks without additional training by using relevant question-answer pairs as references during inference.

To leverage these general-purpose VLMs for dentistry, recent studies~\cite{du2024prompting,wu2025chatios} apply them to dental diagnosis and report promising performance in both diagnostic reasoning and explainability. DentalVLM~\cite{du2024prompting} uses a large-scale bilingual dataset of oral images paired with visual question-answer annotations, covering multiple 2D dental imaging modalities and a broad range of diagnostic tasks. However, these approaches rely on extensive image-text annotations that require domain experts to manually inspect and describe each case, making it difficult to apply these methods across the wide variety of specific dental cases encountered in practice. 
Furthermore, interest continues to grow in vision–language foundation models for medical applications. MedGemma~\cite{sellergren2025medgemma}, for example, introduces a VLM that leverages medical training data, and MMOral~\cite{haotowards} provides a benchmark dataset for general medical analysis with VLMs. However, these models and datasets still primarily rely on specialized clinical imaging such as X-rays or 2D/3D CT/MRI slices, which limits their direct applicability to accessible, smartphone-based dental photographs.
\begin{figure}[t]
    \centering
\includegraphics[width=\columnwidth]{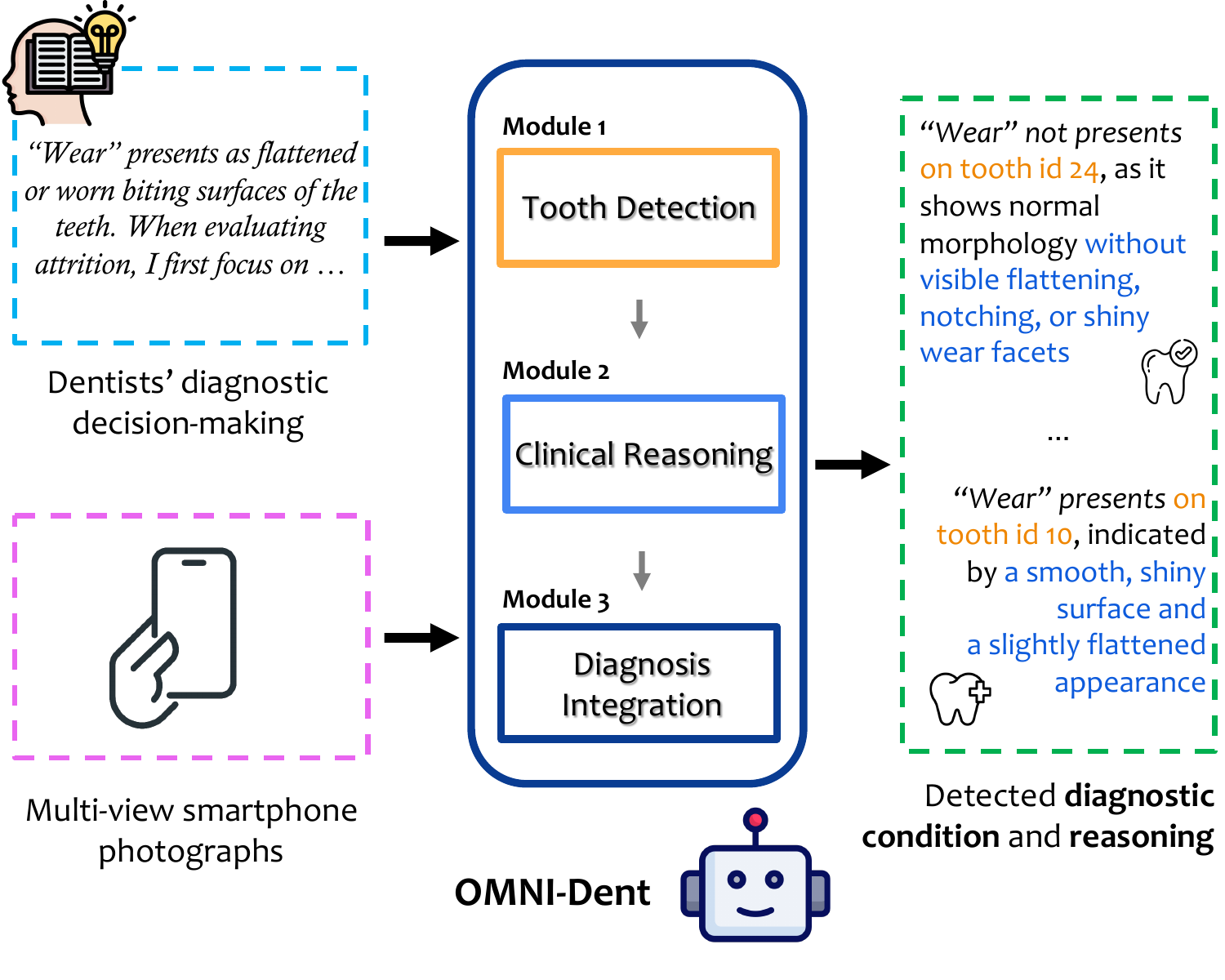}
\caption{
Overview of \model. Replicating dentists' clinical diagnostic reasoning processes, the framework processes multi-view smartphone photographs through tooth detection, clinical reasoning, and diagnosis integration modules. The output of \model\ provides tooth-level diagnostic conditions with corresponding reasoning.
}
    \label{fig:overall}
\end{figure} 

\section{\model}
\label{sec:methods}
In this section, we describe the design and implementation of \model. Sec.~\ref{sec:model_overall} presents an overview of the full diagnostic pipeline and its multi-view input setting. Sec.~\ref{sec:detection} details the tooth-level identification module, which localizes and indexes each tooth from the input views. Sec.~\ref{sec:diagnosis} explains the clinical reasoning module, which replicates expert diagnostic workflows through structured, expert-defined instructions applied to a pretrained VLM. Sec.~\ref{sec:integration} describes the diagnosis integration module, which consolidates per-view predictions into an individual-level diagnostic summary.
\subsection{Framework Overview}~\label{sec:model_overall}
% Figure~X illustrates the overall pipeline of \model. 
Figure~\ref{fig:overall} shows the overview of \model. \model\ takes a set of multi-view smartphone photographs as input, captured from each individual. The inputs consist of three perspectives: (1) a frontal view, (2) an upper occlusal view, and (3) a lower occlusal view, all acquired using commonly available and widely accessible smartphones. These complementary views reveal tooth surfaces that are not fully visible from a single angle and provide sufficient visual diversity to support reliable tooth-level analysis under real-world imaging conditions. 

\model\ comprises three main components: the \textit{tooth  identification module} (Sec.~\ref{sec:detection}), the \textit{clinical reasoning module} (Sec.~\ref{sec:diagnosis}) and the \textit{diagnosis integration module} (Sec.~\ref{sec:integration}). The tooth-level identification module directly localizes each visible tooth across the multi-view inputs and assigns its identifier according to the universal numbering system, thereby establishing precise and standardized indexing.  The clinical reasoning module predicts diagnostic conditions by replicating expert-provided natural-language diagnostic steps. The diagnosis integration module consolidates the overall diagnosis for each tooth. The framework outputs a tooth-level diagnostic condition along with a corresponding reasoning description.

% Sec.~\ref{sec:images} outlines the input images used by \model, and Sec.~\ref{sec:model} presents the methodological overview of the framework and its individual components.

% \subsection{\model}~\label{sec:model}

% \subsection{Multi-View Smartphone Images}~\label{sec:images}

\subsection{Tooth Detection Module}\label{sec:detection}

The tooth detection module aims to localize each tooth in the input images and assign its corresponding universal tooth-numbering identifier. This step is essential for the clinical diagnosis stage (Sec.~\ref{sec:diagnosis}) because it provides explicit tooth-level visual guidance, ensuring that the VLM focuses on clinically relevant evidence at the tooth level rather than relying on broader or ambiguous image regions. We formulate this task as an object detection problem and adopt a YOLOv11-based~\cite{khanam2024yolov11} model to perform joint localization and tooth-ID classification. We train the detector on tooth-level annotations and predicts bounding boxes together with their universal numbering labels, enabling consistent and standardized indexing across images.
\begin{figure*}[t]
    \centering
\includegraphics[height=0.35\textheight]{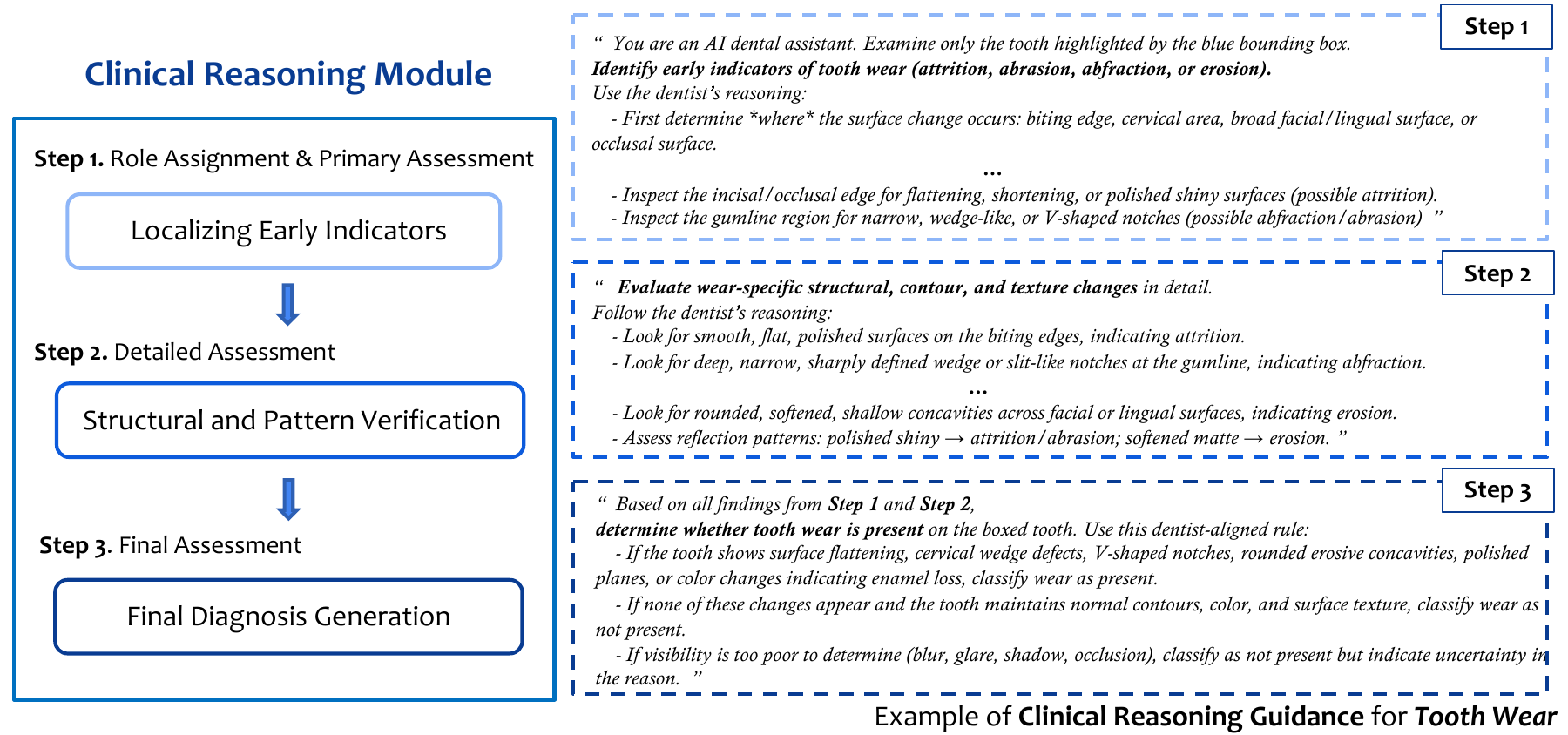}
    \caption{Three-step diagnostic reasoning in the clinical reasoning module (left). For example, the module assesses tooth wear in three stages (right), replicating a dentist's diagnostic process: Step 1 localizes early surface changes; Step 2 examines structural and textural patterns using clinician-guided criteria (e.g., attrition, abfraction/abrasion, erosion); and Step 3 integrates these findings to produce the final diagnosis. }
    \label{fig:clinical_reason}
\end{figure*} 

\subsection{Clinical Reasoning Module} \label{sec:diagnosis}
% \subsection{Clinical Reasoning Integration}

% Add input's information 
\textbf{Visual Input Setting.} After the tooth detection module, \model\ assigns a bounding box and the corresponding universal tooth number to each detected tooth and overlays this information on the image, following the general paradigm of VQA-style queries used in LVLMs (e.g., asking a model to identify the object highlighted by a blue bounding box). This setup specifies the individual tooth region and provides a consistent reference for subsequent processing. By retaining the bounding box while preserving the full image context, the framework enables the model to consider the target tooth in relation to adjacent and contralateral teeth, which is essential for capturing the comparative patterns and structural relationships used in clinical reasoning. 

\paragraph{Clinical Reasoning guidance.}\vspace{-0.5em}
The clinical reasoning module aims to replicate the diagnostic workflow of human dental experts within a VLM pretrained on large-scale, diverse data. To achieve this goal, \model\ introduces expert-defined guidance that encodes the procedural steps, visual criteria, and heuristic assessments used in clinical practice. 

Figure~\ref{fig:clinical_reason} illustrates the three-step diagnostic reasoning process. First, \model\ begins with role assignment and primary assessment, a step that localizes early indicators by specifying the anatomical regions to examine and the morphological or chromatic cues associated with each condition category. Second, \model\ performs structural and pattern verification. This step incorporates heuristic patterns routinely applied by clinicians, including contralateral comparison for symmetry assessment, evaluation of adjacent-tooth relationships to identify cross-tooth patterns, and consistency checks across multiple views of the same tooth. Finally, \model\ generates the final diagnosis by integrating the findings from the previous steps.
% Using these expert-defined instructions, \model\ applies a structured, \textit{multi-step reasoning strategy} to guide a general-purpose VLM in replicating the diagnostic workflow of human dental practitioners. The stepwise instructions encode the procedural steps, visual criteria, and heuristic patterns used in clinical assessment, such as contralateral comparison and evaluation of adjacent-tooth relationships. The sequence consists of three stages: identifying early indicators, evaluating condition-specific structural and chromatic changes, and applying explicit rule-based criteria to assign the final diagnostic label. 
The VLM processes these stages sequentially while retaining intermediate reasoning states, enabling it to emulate clinical diagnostic logic without requiring large-scale dental fine-tuning. For the VLM, we employ the InternVL3~\cite{wang2025internvl3} model, a state-of-the-art vision-language framework pretrained on large-scale image-text data.
% \paragraph{In-Context Learning (ICL)}\vspace{-0.5em}

\subsection{Diagnosis Integration Module}\label{sec:integration}
After generating diagnoses for each image, the diagnosis integration module aggregates the per-tooth predictions across all views to produce a consolidated condition summary for each individual. In addition to assigning condition labels, the module also retains the accompanying reasoning outputs, enabling the framework to provide explanations that remain aligned with the underlying diagnostic logic.

\section{Experiments}
\label{sec:experiments}   % This corresponds to your ``6. Results''

\subsection{Datasets}
We evaluate \model\ using a multi-view dental image dataset, collected at clinical and research settings in the USA and Thailand, with appropriate IRB approval. This paper focuses exclusively on adult participants from the dataset. Each participant contributes three smartphone photographs consisting of frontal, upper occlusal, and lower occlusal views. Licensed dental professionals provide tooth-level diagnostic labels. Given the lower diagnostic accuracy reported for assessing posterior teeth using smartphone-captured images~\cite{estai2022mobile}, we focus on the upper and lower anterior teeth, including the central incisors, lateral incisors, and canines on both sides, for a total of twelve teeth. Under the Universal Numbering System, the corresponding teeth are 6–8, 9–11, 22–24, and 25–27. This design choice ensures consistent visibility across participants and supports reliable evaluation under real-world smartphone imaging conditions. We fully de-identify all images prior to analysis, and data collection and use follow appropriate ethics approval and informed consent. Due to privacy and consent constraints, the dataset remains under restricted access and is not publicly available.

\paragraph{Tooth Detection.}
To identify the exact location of each tooth for the tooth detection module, we manually annotate bounding boxes for every individual tooth along with its FDI tooth number, resulting in 313 annotated images in the dataset. We train the object detection model using a two-step supervised learning strategy: we first pretrain it on 5K open-source tooth-detection images from Roboflow~\cite{dent-ohw72_dataset,intraoral-tooth-numbering-fdi_dataset}, and then fine-tune it on 229 manually annotated smartphone images to align the model with real-world imaging conditions. We evaluate the model on a separate set of 84 manually annotated images from 26 participants, which we do not include during fine-tuning.

\paragraph{Tooth diagnosis.}
To ensure fair evaluation of the full end-to-end framework, we consistently use the same 84 images from the 26 participants for assessing the performance of \model\ and its components. We evaluate \model\ on overall abnormality, which refers to whether each tooth shows any abnormal condition, including tooth wear, uncomplicated crown fracture, or dental caries. We also report the performance for each of the three diagnostic categories individually. 
% Figure~XX summarizes the number of instances for each condition in the evaluation dataset.  

% Dataset Description: Clearly describe all datasets used, including source, size, demographics, and any preprocessing
% Ethics Approval: State whether ethics approval/IRB was obtained and any relevant approval numbers
% Data Availability: Indicate whether data is publicly available or under restricted access
% Limitations: Discuss dataset limitations, potential biases, and generalization concerns
% Patient Privacy: Confirm all data is de-identified and privacy-protected

\subsection{Evaluation Metrics}
To evaluate tooth localization in the \textit{tooth detection module}, we measure multi-class object detection performance, covering both bounding-box localization and tooth ID classification. For the \textit{overall diagnosis} evaluation of \model, we report tooth-level precision, recall, and F1-score to quantify how well the model's predictions align with expert diagnoses across the assessed condition categories.

\subsection{Baselines}
To assess the contribution of each component in \model, we compare the framework with three baseline settings while keeping the same VLM across all experiments. In the first baseline, the VLM analyzes cropped single-tooth images without any clinical reasoning guidance (Exp-1). In the second baseline, the model receives full images without tooth-level localization cues or reasoning guidance (Exp-2). In the third baseline, the model processes full images with a single-tooth bounding box but still does not receive any clinical reasoning prompts (Exp-3). These baselines allow us to isolate the effects of visual input configuration and clinical reasoning guidance, enabling a controlled comparison with the full \model\ setup. For all baseline conditions, we use a general VQA-style question that omits clinical reasoning cues and asks the model to determine whether specific diagnostic categories (e.g., dental caries or tooth wear) are present based solely on its overall visual impression of the individual tooth.

\subsection{Implementation Details}
We conduct all experiments for \model\ on a GPU server equipped with NVIDIA A100-SXM4-40GB GPUs, using the same hardware configuration for both tooth detection inference and clinical reasoning. For the clinical reasoning module, we employ the InternVL3 model with approximately 14 billion parameters as the VLM backbone. We use the model in its pretrained form without any task-specific fine-tuning. For the tooth detection module, we train a YOLOv11-based~\cite{khanam2024yolov11} detector using supervised learning. We train the model using the Adam optimizer with an initial learning rate of 1e-3 for pretraining and 1e-4 for fine-tuning, a batch size of 48, and a total of 40 training epochs. The loss function follows the standard YOLOv11 formulation~\cite{khanam2024yolov11}, incorporating bounding-box regression, objectness prediction, and multi-class classification terms. We pretrain the detector on open-source datasets and fine-tune it on manually annotated images from the Minnesota State Fair dataset to better reflect real-world smartphone imaging conditions.

\section{Results}
\label{sec:results}   % This corresponds to your ``6. Results''
\begin{table*}[t]
\centering
\resizebox{\textwidth}{!}{
\begin{tabular}{l | l | c | c | c| c c c  | c c | c}
\toprule
\textbf{Diagnosis Category} & \textbf{Experiment} &
\textbf{Visual Input Setting} & \textbf{Clinical Reasoning Guidance} &
\textbf{Actual Positive} & \textbf{TP} & \textbf{FP} & \textbf{FN} &
\textbf{Precision} & \textbf{Recall} & \textbf{F1-score} \\
\toprule[1pt]\midrule[0.3pt]

\multirow{4}{*}{Overall Abnormality}
& Exp-1 & Cropped single-tooth image & No & \multirow{4}{*}{271} & 89 & 25 & 182  & 0.78 &  0.32 & 0.46\\
& Exp-2 & Full image & No &&  155 & 22 &  116 & \textbf{0.87} &  \underline{0.57} & \underline{0.69}\\
& Exp-3 & Full image + single tooth bounding box & No &&   141 &   33  & 130 & \underline{0.81}  & 0.52 & 0.63 \\
& \model\ & Full image + single tooth bounding box & Yes &&  264 & 62 &   7 &  0.80 &  \textbf{0.97} & \textbf{0.88}\\
\toprule[1pt]\midrule[0.3pt]
\multirow{4}{*}{Wear}
& Exp-1 & Cropped single-tooth image & No & \multirow{4}{*}{215}& 38 & 29 & 177 & 0.56 & 0.17 & 0.26 \\
& Exp-2 & Full image & No && 86 & 38 & 129  & \underline{0.69} & \underline{0.40} & \underline{0.50} \\
& Exp-3 & Full image + single tooth bounding box & No && 68
 & 29 & 147 & \textbf{0.70} & 0.31 & 0.43 \\
& \model\ & Full image + single tooth bounding box & Yes && 206 & 115 & 9 & 
0.64 & \textbf{0.95} & \textbf{0.76} \\
\midrule
\multirow{4}{*}{\shortstack{Uncomplicated \\ Crown Fracture}}
& Exp-1 & Cropped single-tooth image & No & \multirow{5}{*}{43} & 0 & 3 & 43 & 0.00 & 0.00 & 0.00 \\
& Exp-2 & Full image & No && 1 & 1 & 42 & \textbf{0.50} & \underline{0.02} & \underline{0.04} \\
& Exp-3 & Full image + single tooth bounding box & No && 0 & 3 & 43 & 0.00 & 0.00 & 0.00 \\
& \model\ & Full image + single tooth bounding box & Yes && 27 & 159 & 16 & 
\underline{0.14} & \textbf{0.62} & \textbf{0.23} \\
% & \model\ + ICL & Full image + single tooth bounding box & Yes && 16  &  37& 27  & 
% \underline{0.30} & \underline{0.37} & \textbf{0.33} \\
\midrule
\multirow{4}{*}{Dental Caries}
& Exp-1 & Cropped single-tooth image & No & \multirow{5}{*}{16}& 5 & 41 & 11 & 0.10 & 0.31  & 0.16  \\
& Exp-2 & Full image & No && 7 & 84 & 9 & 0.07  & 0.43  & 0.13 \\
& Exp-3 & Full image + single tooth bounding box & No && 11
 & 82 & 5 & \underline{0.11} & \textbf{0.68} & \underline{0.20} \\
& \model\ & Full image + single tooth bounding box & Yes && 11 & 59 & 5  & 
\textbf{0.15} & \textbf{0.68} & \textbf{0.25} \\
% & \model\ + ICL & Full image + single tooth bounding box & Yes && 1 & 3 & 15  & 
% 0.25 & 0.06 & 0.10 \\

\bottomrule
\end{tabular}
}
\caption{
Evaluation results across four experimental settings for overall abnormality detection and three diagnostic categories (tooth wear, uncomplicated crown fracture, and dental caries). Bold values indicate the best performance, and underlined values indicate the second-best.
}
\label{tab:overall}
\end{table*}

\subsection{Framework Evaluation}
Table~\ref{fig:overall} reports the evaluation results for \model. We first examine overall abnormality, which indicates whether each tooth shows any condition such as tooth wear, uncomplicated crown fracture, or dental caries. We then present observations for each diagnostic category individually.

\paragraph{Overall abnormality.}

The overall abnormality category includes tooth wear, uncomplicated crown fracture, and dental caries, totaling 271 positive cases. This setting best reflects \model's role as an early-stage screening tool.

Across the baselines, Exp-1 produces the lowest performance because the cropped single-tooth image removes surrounding anatomical context and prevents the VLM from leveraging cross-tooth relationships. Exp-2 and Exp-3 achieve similar performance levels: Exp-2 gains broader visual cues from the full image, and Exp-3 stabilizes attention by highlighting the target tooth, but both settings still lack the diagnostic reasoning needed to detect subtle abnormalities. As a result, the best baseline score appears in Exp-2 with an F1-score of 0.69. In contrast, \model\ reaches an F1-score of 0.88, outperforming all baselines by a substantial margin. Notably, \model\ achieves this improvement without any additional training that relies on large-scale annotations, highlighting the effectiveness of explicit visual guidance and clinical reasoning in enabling a general-purpose VLM to perform reliable abnormality detection.

\paragraph{Tooth Wear.}

Tooth wear includes 215 positive cases, the largest among the diagnostic categories. Exp-1 produces the lowest performance because the cropped single-tooth view removes adjacent-tooth context and prevents the VLM from comparing surface patterns across neighboring teeth. This limitation results in an F1-score far below that of \model, which improves performance by 0.5. Exp-2 and Exp-3 achieve higher scores than Exp-1 by providing full-image context or explicit localization, yet both settings still fall short of \model\ by a substantial margin, with F1-score differences of 0.26–0.29. These gaps highlight that, even under the same VLM, reliable tooth-wear detection requires both explicit visual guidance and clinical reasoning cues. 

\paragraph{Uncomplicated Crown Fracture.}

Uncomplicated crown fracture includes 43 positive cases and typically appears as subtle enamel fractures or faint craze lines involving dentin~\cite{antipoviene2021traumatic}. These features challenge visual interpretation in smartphone photographs, especially when images are blurry or cues are minimal. These characteristics make uncomplicated crown fractures particularly difficult for a general-purpose VLM to recognize in smartphone photographs, and all baseline settings perform extremely poorly as a result. Exp-1, Exp-2, and Exp-3 all yield near-zero F1-scores (with the best baseline reaching only 0.04). In contrast, \model\ attains an F1-score of 0.23, representing a substantial improvement over all baselines and highlighting the essential role of clinical reasoning guidance in detecting fine-grained structural defects. Identifying uncomplicated crown fractures requires deliberate evaluation of enamel continuity across multiple views, a process that general-purpose VLMs cannot perform without explicit reasoning guidance.

\paragraph{Dental Caries.}

Dental caries includes 16 positive cases, which makes the evaluation sensitive to small prediction errors. Exp-1 relies solely on a cropped view and often confuses staining or food debris with dental caries, which lowers precision. Exp-2 adds global context but provides no localization, causing the model to attend to irrelevant regions and further degrade precision. Exp-3 localizes the target tooth and improves recall, yet the VLM still struggles to differentiate discoloration from true lesions without diagnostic reasoning. While both \model\ and the baselines show lower precision than recall for dental caries detection, \model\ improves recall to 0.68 and attains an F1-score of 0.25, outperforming the baseline methods, which remain substantially lower in both recall and F1-score.

\paragraph{Evaluation Summary.}

Across all diagnostic categories, \model\ consistently outperforms the baseline configurations by a large margin. Exp-1, Exp-2, and Exp-3 each reveal different limitations of a general-purpose VLM, such as restricted context, ambiguous localization, or the absence of diagnostic reasoning, which lead to low recall and unstable performance. In contrast, \model\ combines explicit visual guidance with clinical reasoning and achieves substantial gains without any additional annotation-heavy training.  These results demonstrate that clinical reasoning guidance is essential for enabling a VLM to reliably interpret subtle dental cues in real-world smartphone images. Furthermore, we provide an analysis of common failure patterns in the Section~\ref{sec:failure}.
\begin{table}[t]
\centering
\resizebox{\columnwidth}{!}{
\begin{tabular}{l | l | c c c}
\toprule
\textbf{Category} & \textbf{Experiment} & \textbf{Precision} & \textbf{Recall} & \textbf{F1-score} \\
\midrule
\multirow{2}{*}{\shortstack{Uncomplicated \\ Crown Fracture}}
& \model & 0.14 & \textbf{0.62} & 0.23 \\
& \model + ICL & \textbf{0.30} & 0.37 & \textbf{0.33} \\
\midrule
\multirow{2}{*}{Dental Caries}
& \model & 0.15 & \textbf{0.68} & \textbf{0.25} \\
& \model + ICL & \textbf{0.25} & 0.06 & 0.10 \\
\bottomrule
\end{tabular}
}
\caption{
Comparison of \model\ and \model\,+\,ICL on two diagnostic categories. Bold indicates the best score.
}
\label{tab:icl_results}
\end{table}

\subsection{In-Context Learning (ICL) Capabilities}

In-context learning (ICL) allows VLMs to adapt to domain-specific tasks without additional training by referencing a few example question–answer pairs during inference.  In this work, we further examine whether \model\ benefits from ICL when diagnosing subtle dental conditions. For each patient, the dataset includes paired views of the same tooth (e.g., frontal and occlusal). We construct an ICL prompt using two paired images that include the tooth bounding box and the corresponding expert-provided diagnosis. At inference time, we provide two such reference pairs (four images in total) and evaluate the model's predictions on a separate image of the same tooth.

We evaluate ICL on uncomplicated crown fracture and dental caries, the two categories with the most subtle and variable visual appearances. Table~\ref{tab:icl_results} reports the quantitative results. ICL yields a clear improvement for uncomplicated crown fracture, raising the F1-score from 0.23 to 0.33. Because fracture lines exhibit relatively consistent structural cues across views, the reference examples help the VLM better distinguish these defects from normal surface texture. In contrast, ICL does not improve performance for dental caries. Caries exhibits highly variable visual patterns and often resembles staining or food debris. With only two paired ICL examples, the model lacks sufficient information to resolve these ambiguities, and the additional examples may even narrow or distort the model's internal decision patterns.

% These observations suggest that ICL is effective when the diagnostic category exhibits stable and repeatable visual patterns, but less effective for conditions with large intra-class variability or strong overlap with non-pathological appearances.

\begin{table}[t]
\centering
\resizebox{\columnwidth}{!}{
\begin{tabular}{llccc}
\toprule
\textbf{Pretrained} & \textbf{fine-tuned} & \textbf{Precision} & \textbf{Recall} & \textbf{F1-score} \\
\midrule
Yes & No & 0.80 & 0.79 & 0.80 \\
 Yes & Yes & \textbf{0.96} & \textbf{0.89} & \textbf{0.92} \\
\bottomrule

\end{tabular}
}
\caption{Tooth detection performance before and after fine-tuning on smartphone images.Bold indicates the best score.}
\label{tab:tooth_detection}
\end{table} 
\subsection{Tooth detection evaluation} 
We evaluate the performance of the tooth detection module by examining the impact of the two-step training strategy. We first pretrain the detector on open-source tooth-detection images, each containing bounding-box annotations for individual teeth. We then fine-tune the model on manually annotated images from the dataset to adapt the detector to real-world smartphone imaging characteristics. Table~\ref{tab:tooth_detection}
summarizes the detection results. The pretrained model achieves a precision of 0.80, a recall of 0.79, and an F1-score of 0.80, indicating reasonable transferability from general-purpose tooth images. After fine-tuning on smartphone-specific images, performance improves substantially, reaching a precision of 0.96, a recall of 0.89, and an F1-score of 0.92. These results demonstrate that domain adaptation through fine-tuning is crucial for reliable tooth localization and ID classification in smartphone photographs.
\section{Discussion}
\label{sec:discussion}   % This corresponds to your ``7. Discussion''
% Critical section: Clinical implications, deployment considerations, limitations, failure cases, and future work.
\subsection{Failure Cases Analysis}\label{sec:failure}

We examine the failure cases of \model\ to understand the diagnostic errors that arise in smartphone images. Because \model\ produces explicit tooth-level reasoning for each prediction, we can directly inspect the visual cues and decision steps that lead to misclassification. This interpretability enables a more detailed and clinically meaningful failure analysis than prediction-only models.  

For tooth wear, errors often occur when the occlusal surface is not clearly visible. In these situations, the model relies on secondary cues such as the incisal edge, where mild flattening or reflections can resemble attrition and lead to overdiagnosis. For uncomplicated crown fracture, false positives frequently arise from normal surface irregularities or reflections that mimic faint enamel fractures. Illumination and viewing angles in smartphone photographs can exaggerate these cues, making trauma detection particularly difficult. For dental caries, staining or food debris often creates dark regions that resemble early carious lesions. Limited contextual information and the small number of caries cases increase the likelihood of misinterpreting these regions as pathological. 

% Overall, these errors reflect challenges inherent to uncontrolled smartphone imaging rather than limitations of the reasoning module itself. 

\subsection{Clinical Implications}
\model\ suggests that a general-purpose VLM, when guided by expert-defined diagnostic reasoning, can approximate key aspects of clinical decision-making without requiring labor-intensive fine-tuning on large, expert-annotated dental datasets. The strong recall achieved across all diagnostic categories is particularly meaningful for an assistive tool intended to encourage timely dental care, as it reduces the likelihood of missed abnormalities that could otherwise delay treatment. 

In addition, the stepwise reasoning outputs generated by \model\ provide interpretable justifications aligned with clinical logic, which may support early-career clinicians and offer patients clearer explanations of why a finding warrants professional evaluation. These explicit reasoning traces also contribute to explainable AI by revealing how the model arrives at its conclusions, enabling more targeted feedback and facilitating iterative refinement of the system's diagnostic behavior. Consequently, \model\ has the potential to connect accessible smartphone-based imaging with clinically informed diagnostic guidance, improving access to preliminary oral health assessment for individuals in underserved or marginalized communities who face barriers to timely professional care. 

\subsection{Deployment Considerations}
Considering that \model\ operates on widely accessible smartphone photographs, it supports deployment in settings that lack specialized dental imaging equipment. However, reliable and effective deployment depends on several factors. First, the system guides users to capture images of sufficient quality to ensure reliable tooth localization and diagnosis. Second, privacy protections play a critical role, as smartphone images may include identifiable facial features or metadata, and the system handles this information securely. Finally, although the system demonstrates strong recall, it does not replace professional diagnosis. Appropriate human oversight remains necessary, and deployment emphasizes that the system provides preliminary assessments to assist, rather than substitute, clinical judgment.

\subsection{Limitations}
Despite the promising findings, \model\ exhibits several limitations. First, the evaluation dataset includes limited size and demographic diversity, which may constrain generalizability across broader populations and imaging conditions. Certain conditions, such as decay, appear infrequently in the dataset (e.g., 15 instances), reducing the reliability of condition-specific performance estimates. Second, improving diagnostic performance (Sec.~\ref{sec:diagnosis}) remains challenging for conditions whose visual presentation is subtle or easily confused with normal variations (e.g., uncomplicated crown fracture). These cases often lack distinct visual boundaries and may require additional contextual or non-visual information for reliable differentiation. Third, the current framework performs limited reasoning during the aggregation of multi-view predictions. The diagnosis integration module (Sec.~\ref{sec:integration}) does not yet incorporate view-specific contextual descriptions or confidence-based weighting, which may lead to less reliable combined diagnoses across views.

\subsection{Future Work}
Future work will expand the diagnostic scope of \model\ and improve its robustness under real-world smartphone imaging conditions. Extending the set of diagnostic categories, including restorations and other clinically relevant conditions, and improving the reliability of the tooth detection module (Sec.~\ref{sec:detection}) across diverse visual presentations remain important directions. Future work will also strengthen the clinical reasoning module (Sec.~\ref{sec:diagnosis}). Exploring medical-domain VLMs, integrating external knowledge sources, and enhancing reasoning through in-context learning (ICL) may improve performance for conditions with subtle or ambiguous cues. In addition, refining multi-view aggregation with view-specific context or confidence-based weighting in the diagnosis integration module (Sec.~\ref{sec:integration}) may further improve diagnostic consistency. Finally, prospective evaluations in clinical and community settings will be essential for assessing usability, user trust, and the broader impact of \model, particularly for populations with limited access to professional care.

\section{Conclusion}
\label{sec:conclusion}   % This corresponds to your ``8. Conclusion''

This paper presents \model, an explainable and data-efficient framework for automated dental diagnosis using multi-view smartphone photographs. By combining tooth-level detection with expert-defined clinical reasoning instructions, the framework enables a general-purpose VLM to emulate structured diagnostic workflows without large-scale dental fine-tuning. Experimental results show that \model\ consistently outperforms baseline configurations across all evaluated conditions, which is essential for an assistive system intended to encourage timely professional evaluation. Through its integration of clinically grounded reasoning, transparent outputs, and accessible imaging requirements, \model\ contributes toward more interpretable, reliable, and widely accessible dental assessment that may benefit individuals with limited access to direct dental care.

{\small
\bibliographystyle{ieeenat_fullname}
\bibliography{main}}

@article{rahman2024dental,
  title={Dental clinic deserts in the US: spatial accessibility analysis},
  author={Rahman, Md Shahinoor and Blossom, Jeffrey C and Kawachi, Ichiro and Tipirneni, Renuka and Elani, Hawazin W},
  journal={JAMA Network Open},
  volume={7},
  number={12},
  pages={e2451625--e2451625},
  year={2024},
  publisher={American Medical Association}
}

@article{wang2025application,
  title={Application of machine learning in dentistry: insights, prospects and challenges},
  author={Wang, Lin and Xu, Yanyan and Wang, Weiqian and Lu, Yuanyuan},
  journal={Acta Odontologica Scandinavica},
  volume={84},
  pages={43345},
  year={2025}
}

@article{abbott2025artificial,
  title={Artificial Intelligence Platforms in Dental Caries Detection: A Systematic Review and Meta-Analysis},
  author={Abbott, Lyndon P and Saikia, Ankita and Anthonappa, Robert P},
  journal={Journal of Evidence-Based Dental Practice},
  volume={25},
  number={1},
  pages={102077},
  year={2025},
  publisher={Elsevier}
}

@inproceedings{dan2025artifact,
  title={Artifact Correction in Panoramic Radiographs using Deep De-shadowing},
  author={Dan, Omri and Lilek, Samuel and Hirschhorn, Ariel and Kats, Lazar and Kiryati, Nahum and Mayer, Arnaldo},
  booktitle={Proceedings of the IEEE/CVF International Conference on Computer Vision},
  pages={1008--1016},
  year={2025}
}

@article{santos2025unique,
  title={A unique AI-based tool for automated segmentation of pulp cavity structures in maxillary premolars on CBCT},
  author={Santos-Junior, Airton Oliveira and Fontenele, Rocharles Cavalcante and Neves, Frederico Sampaio and Ali, Saleem and Jacobs, Reinhilde and Tanomaru-Filho, M{\'a}rio},
  journal={Scientific Reports},
  volume={15},
  number={1},
  pages={5509},
  year={2025},
  publisher={Nature Publishing Group UK London}
}

@article{zhang2022development,
  title={Development and evaluation of deep learning for screening dental caries from oral photographs},
  author={Zhang, Xuan and Liang, Yuan and Li, Wen and Liu, Chao and Gu, Deao and Sun, Weibin and Miao, Leiying},
  journal={Oral diseases},
  volume={28},
  number={1},
  pages={173--181},
  year={2022},
  publisher={Wiley Online Library}
}

@inproceedings{du2024prompting,
  title={Prompting Vision-Language Models for Dental Notation Aware Abnormality Detection},
  author={Du, Chenlin and Chen, Xiaoxuan and Wang, Jingyi and Wang, Junjie and Li, Zhongsen and Zhang, Zongjiu and Lao, Qicheng},
  booktitle={International Conference on Medical Image Computing and Computer-Assisted Intervention},
  pages={687--697},
  year={2024},
  organization={Springer}
}

@article{wu2025chatios,
  title={ChatIOS: Improving automatic 3-dimensional tooth segmentation via GPT-4V and multimodal pre-training},
  author={Wu, Yongjia and Zhang, Yun and Wu, Yange and Zheng, Qianhan and Li, Xiaojun and Chen, Xuepeng},
  journal={Journal of Dentistry},
  volume={157},
  pages={105755},
  year={2025},
  publisher={Elsevier}
}

@inproceedings{chen2024internvl,
  title={Internvl: Scaling up vision foundation models and aligning for generic visual-linguistic tasks},
  author={Chen, Zhe and Wu, Jiannan and Wang, Wenhai and Su, Weijie and Chen, Guo and Xing, Sen and Zhong, Muyan and Zhang, Qinglong and Zhu, Xizhou and Lu, Lewei and others},
  booktitle={Proceedings of the IEEE/CVF conference on computer vision and pattern recognition},
  pages={24185--24198},
  year={2024}
}

@article{wang2025internvl3,
  title={Internvl3. 5: Advancing open-source multimodal models in versatility, reasoning, and efficiency},
  author={Wang, Weiyun and Gao, Zhangwei and Gu, Lixin and Pu, Hengjun and Cui, Long and Wei, Xingguang and Liu, Zhaoyang and Jing, Linglin and Ye, Shenglong and Shao, Jie and others},
  journal={arXiv preprint arXiv:2508.18265},
  year={2025}
}

@article{lu2024ovis,
  title={Ovis: Structural embedding alignment for multimodal large language model},
  author={Lu, Shiyin and Li, Yang and Chen, Qing-Guo and Xu, Zhao and Luo, Weihua and Zhang, Kaifu and Ye, Han-Jia},
  journal={arXiv preprint arXiv:2405.20797},
  year={2024}
}

@article{bai2023qwen,
  title={Qwen technical report},
  author={Bai, Jinze and Bai, Shuai and Chu, Yunfei and Cui, Zeyu and Dang, Kai and Deng, Xiaodong and Fan, Yang and Ge, Wenbin and Han, Yu and Huang, Fei and others},
  journal={arXiv preprint arXiv:2309.16609},
  year={2023}
}

@article{sellergren2025medgemma,
  title={Medgemma technical report},
  author={Sellergren, Andrew and Kazemzadeh, Sahar and Jaroensri, Tiam and Kiraly, Atilla and Traverse, Madeleine and Kohlberger, Timo and Xu, Shawn and Jamil, Fayaz and Hughes, C{\'\i}an and Lau, Charles and others},
  journal={arXiv preprint arXiv:2507.05201},
  year={2025}
}

@article{achiam2023gpt,
  title={Gpt-4 technical report},
  author={Achiam, Josh and Adler, Steven and Agarwal, Sandhini and Ahmad, Lama and Akkaya, Ilge and Aleman, Florencia Leoni and Almeida, Diogo and Altenschmidt, Janko and Altman, Sam and Anadkat, Shyamal and others},
  journal={arXiv preprint arXiv:2303.08774},
  year={2023}
}

@inproceedings{haotowards,
  title={Towards Better Dental AI: A Multimodal Benchmark and Instruction Dataset for Panoramic X-ray Analysis},
  author={Hao, Jing and Fan, Yuxuan and Sun, Yanpeng and Guo, Kaixin and Lizhuo, Lin and Yang, Jinrong and Ai, Qiyong Hemis and Wong, Lun M and Tang, Hao and Hung, Kuo Feng},
  booktitle={The Thirty-ninth Annual Conference on Neural Information Processing Systems Datasets and Benchmarks Track}
}

@online{intraoral-tooth-numbering-fdi_dataset,
  title        = {Intraoral Tooth Numbering FDI Dataset},
  author       = {DentalMate6v},
  organization = {Roboflow Universe},
  year         = {2024},
  month        = jul,
  url          = {https://universe.roboflow.com/dentalmate6v/intraoral-tooth-numbering-fdi},
  note         = {Open source dataset, visited on 2025-12-02}
}

@online{dent-ohw72_dataset,
  title        = {dent Dataset},
  author       = {dental},
  organization = {Roboflow Universe},
  year         = {2023},
  month        = oct,
  url          = {https://universe.roboflow.com/dental-cqxpi/dent-ohw72},
  note         = {Open source dataset, visited on 2025-12-02}
}

@article{khanam2024yolov11,
  title={YOLOV11: AN OVERVIEW OF THE KEY ARCHITECTURAL ENHANCEMENTS},
  author={Khanam, Rahima and Hussain, Muhammad},
  journal={arXiv preprint arXiv:2410.17725},
  year={2024}
}

@article{peres2019oral,
  title={Oral diseases: a global public health challenge},
  author={Peres, Marco A and Macpherson, Lorna MD and Weyant, Robert J and Daly, Bl{\'a}naid and Venturelli, Renato and Mathur, Manu R and Listl, Stefan and Celeste, Roger Keller and Guarnizo-Herre{\~n}o, Carol C and Kearns, Cristin and others},
  journal={The Lancet},
  volume={394},
  number={10194},
  pages={249--260},
  year={2019},
  publisher={Elsevier}
}

@inproceedings{azimi2025teledentistry,
  title={Teledentistry Improves Access to Oral Care: A Cluster Randomised Controlled Trial},
  author={Azimi, Somayyeh and Bennamoun, Basheer and Mehdizadeh, Maryam and Vignarajan, Janardhan and Xiao, Di and Huang, Boyen and Spallek, Heiko and Irving, Michelle and Kruger, Estie and Tennant, Marc and others},
  booktitle={Healthcare},
  volume={13},
  number={18},
  pages={2282},
  year={2025},
  organization={MDPI}
}

@article{huang2024mobile,
  title={Mobile health assessment of traumatic dental injuries using smartphone-acquired photographs: a multicenter diagnostic accuracy study},
  author={Huang, Boyen and Estai, Mohamed and Pungchanchaikul, Patimaporn and Quick, Karin and Ranjitkar, Sarbin and Fashingbauer, Emily and Askar, Abdirahim and Wang, Josiah and Diefalla, Fatma and Shenouda, Margaret and others},
  journal={Telemedicine and e-Health},
  volume={30},
  number={10},
  pages={2592--2600},
  year={2024},
  publisher={Mary Ann Liebert, Inc., publishers 140 Huguenot Street, 3rd Floor New~…}
}

@article{estai2017comparison,
  title={Comparison of a smartphone-based photographic method with face-to-face caries assessment: a mobile teledentistry model},
  author={Estai, Mohamed and Kanagasingam, Yogesan and Huang, Boyen and Shiikha, Julia and Kruger, Estie and Bunt, Stuart and Tennant, Marc},
  journal={Telemedicine and e-Health},
  volume={23},
  number={5},
  pages={435--440},
  year={2017},
  publisher={Mary Ann Liebert, Inc. 140 Huguenot Street, 3rd Floor New Rochelle, NY 10801 USA}
}

@article{estai2022mobile,
  title={Mobile photographic screening for dental caries in children: diagnostic performance compared to unaided visual dental examination},
  author={Estai, Mohamed and Kanagasingam, Yogesan and Mehdizadeh, Maryam and Vignarajan, Janardhan and Norman, Richard and Huang, Boyen and Spallek, Heiko and Irving, Michelle and Arora, Amit and Kruger, Estie and others},
  journal={Journal of Public Health Dentistry},
  volume={82},
  number={2},
  pages={166--175},
  year={2022},
  publisher={Wiley Online Library}
}

@article{estai2016efficacy,
  title={The efficacy of remote screening for dental caries by mid-level dental providers using a mobile teledentistry model},
  author={Estai, Mohamed and Kanagasingam, Yogesan and Huang, Boyen and Checker, Hellen and Steele, Lesley and Kruger, Estie and Tennant, Marc},
  journal={Community Dentistry and Oral Epidemiology},
  volume={44},
  number={5},
  pages={435--441},
  year={2016},
  publisher={Wiley Online Library}
}

@article{antipoviene2021traumatic,
  title={Traumatic dental injuries, treatment, and complications in children and adolescents: a register-based study},
  author={Antipovien{\.e}, Aust{\.e} and Narbutait{\.e}, Julija and Virtanen, Jorma I},
  journal={European journal of dentistry},
  volume={15},
  number={03},
  pages={557--562},
  year={2021},
  publisher={Thieme Medical and Scientific Publishers Pvt. Ltd.}
}

@article{estai2016proof,
  title={A proof-of-concept evaluation of a cloud-based store-and-forward telemedicine app for screening for oral diseases},
  author={Estai, Mohamed and Kanagasingam, Yogesan and Xiao, Di and Vignarajan, Janardhan and Huang, Boyan and Kruger, Estie and Tennant, Marc},
  journal={Journal of telemedicine and telecare},
  volume={22},
  number={6},
  pages={319--325},
  year={2016},
  publisher={SAGE Publications Sage UK: London, England}
}

@article{schultz2025review,
  title={In-Review: Perspectives of Front-Line Clinicians and Remote Reviewers on Smartphone-Based Photography for Assessing Traumatic Dental Injuries: A Qualitative Study},
  author={Schultz, Emily C and Huang, Boyen and Shenouda, Margaret and Estai, Mohamed and Ranjitkar, Sarbin and Louie, Jeffrey P and Pungchanchaikul, Patimaporn},
  year={2025},
  publisher={Preprint}
}
\end{document}